# Bayesian Control for Concentrating Mixed Nuclear Waste

Robert L. Welch
Gensym Corporation
Boulder, Colorado 80303
bwelch@gensym.com

Clayton Smith
University of Denver Research Institute
Denver, Colorado 80208
clasmith@du.edu

**Abstract**

A control algorithm for batch processing of mixed waste is proposed based on conditional Gaussian Bayesian networks. The network is compiled during batch staging for real-time response to sensor input.

## INTRODUCTION

Mixed waste remediation is an example of an industrial operation that transforms input of low quality, high variability and uncertain composition into consistent and high quality output. Moreover, control of many of the processing steps is dependent on an accurate knowledge of the composition of the material input. When control is not optimized, product quality may suffer, processing time may increase maintenance and repairs may be more frequent, and severe damage to equipment is possible. Yet, obtaining accurate knowledge or estimates of the input stream composition may be impossible or time consuming, may reduce throughput, and may increase costs. With this uncertainty, optimal control must balance the costs of obtaining more accurate knowledge about the input stream against costs of repair, maintenance, associated shutdowns, and, generally, reduced throughput.

Here, a Bayesian controller is used in real time to manage this tradeoff. A Bayesian network predicts the quality of the input stream. Sensor input is used to derive a posterior distribution of the input composition. From the latter, the optimal control setting is computed.

Complexity Workarounds

Although Bayesian networks provide superior representation and inference capabilities, it is also known that "solving" a Bayesian network is NP-hard (Cooper [1990]). Hence use of a Bayesian network for real-time control is problematic.

A common work-around to Bayesian network complexity is to use approximate solutions often in the form of "anytime" algorithms (Dean and Boddy [1994], Horvitz [1987], Ibarguengoytia et al [1998], Horsch and Poole [1998]). Here we demonstrate a different approach, one that has largely been ignored within the uncertainty in AI community, based on exploiting differences in the update frequency of sensor variables. A large complex Bayesian network is needed for assessment of the joint distribution of all system variables relevant to the real time decision problem. Yet only a small portion of the network is needed at run time for the most frequent control decisions. This reduced distribution corresponds to a conditional sub-network of the original Bayesian network, which can be evaluated, in real time, using standard exact solution methods.

In this paper we apply this approach to the control of a process having both batch and continuous components. The staging of the batches provides ample time to update the larger Bayesian network and compile the smaller sub-network used for control of the continuous processing of the batch.

## 1. THE SETTING: WASTE REMEDIATION

We work with an example of the cleanup of buried nuclear waste. A burial site has received nuclear waste of various types in a random fashion over a number of years. The last addition was several decades ago, and by now container damage and leakage have resulted in an undetermined level of soil contamination. The task (see figure 1) is to retrieve the waste and soil separately into large boxes (100 cu. ft.), then to assay each box for transuranic (TRU) contamination. This TRU assay uses a relatively slow, but relatively precise, neutron emission method. The boxes that exceed an assay threshold for TRU content are sent to further processing, those below the threshold are returned to the pit.

The high waste boxes move to vitrification; the high soil boxes are sent to a segmented gate, soil sorter (γ-sort) to concentrate the contamination into a reduced volume. This device works by assaying small volumes (0.1 cu. ft.) of soil using a gamma detector, which provides a very rapid, but much less precise method. As before, high level portions are sent to vitrification, low level portions are returned to the pit, after verification by the relatively precise neutron assay.

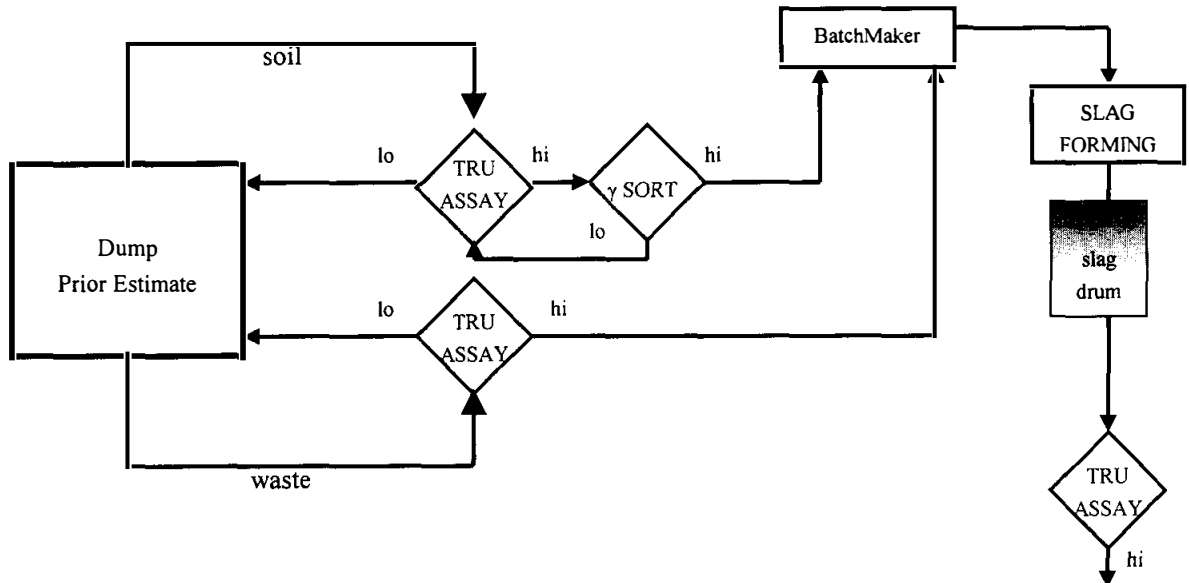

Figure 1. Material Flow in Nuclear Waste Retrieval and Processing

The objective of the Bayesian controller is to establish a sorting algorithm, which will optimize the operation of the soil sorter, under conditions of very high uncertainty.

The radioisotopes that are the source of gamma are not the long lived TRU isotopes. Instead they accompany the TRU in proportions that vary with the type of waste. If the ratio of the gamma emitter to the neutron emitting TRU were known, then the bias in the sensor could be compensated. However, depending on the origin of the TRU-waste, the ratio of gamma emissions to neutron emissions can vary by factors of over $10^4$ among the types of wastes, and by a factor of 10 within a waste type.

In addition, the amount of contamination in any portion of soil under assay will vary significantly from sample to sample. A small speck of waste with a high gamma, low TRU contamination will provide a stronger signal, than a large bit of low gamma, high TRU waste. The specifications, which must be met, are based on the slow, neutron emission analysis for TRU. These considerations make the gamma signal difficult to interpret. If the waste type where known, then a higher gamma indicates a greater TRU contamination. But with mixed waste, a higher gamma may instead indicate presence of a waste type that is a high gamma emitter but has low TRU contamination.

Yet, operational and economic considerations are compelling for using the gamma emissions sensor for sorting. The primary cost considerations are the volume of the slag produced and the throughput of the sorter.

Although the waste composition and the extent of the spread of contamination are not known with certainty, an imperfect prior estimate can be obtained from several sources: dump records, waste composition logs, processing records, non intrusive surveys of the waste site, and the assays of process streams which are anterior and posterior to the sorter.

Therefore, in light of uncertainty about the composition of waste dispersed into the soil, we have considered the use of optimal Bayesian control to minimize the volume of material, which must be vitrified. Using Bayesian methods, assay and sensor readings lead to revised beliefs in the type of waste in the soil sample, thus to better control decision.

The same economic considerations that led to the selection of the gamma sensor for its greater throughput also require that the sorter algorithm be fast, with a sub-second response between reading and control actuation.

## 2. BAYESIAN NETWORK BASED SORTER CONTROL

Although uncertainty and improved knowledge of the composition of the waste and soil streams is important at several stages in the remediation process, we concentrate in this paper on the segmented sorter. The main characteristics of this process control problem are the following:

1. A multi criteria objective to achieve high throughput and low slag volume subject to maximum contamination constraints on soil returned to the pit and minimal contamination of slag drums. The parameters are specified contractually.

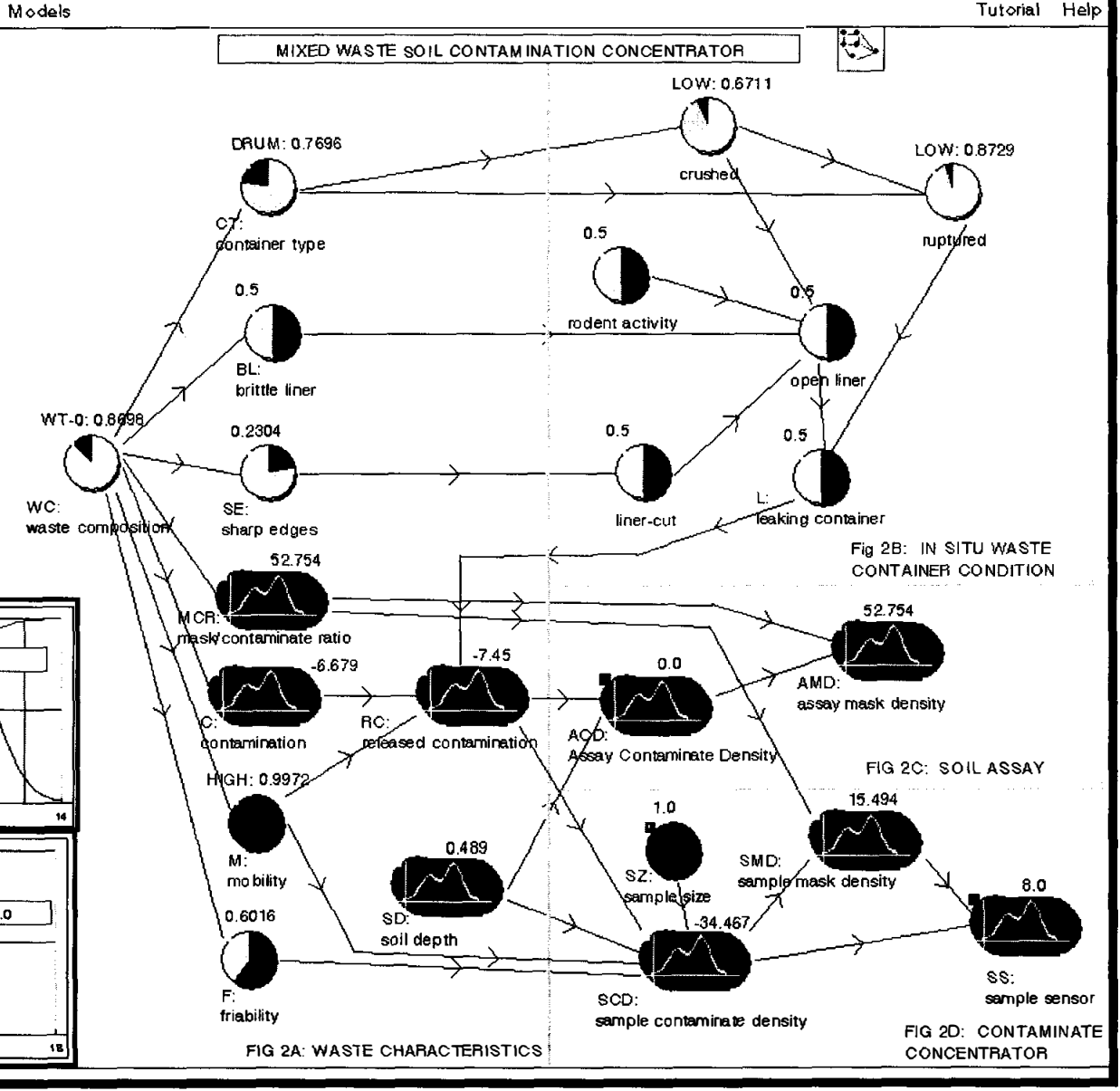
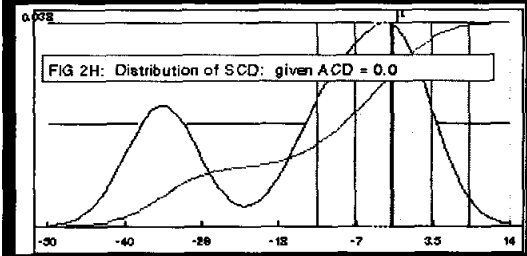
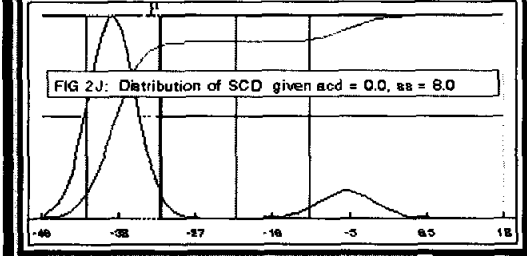

Figure 2. Displayed is the full Bayesian network for the contaminate concentrator model. Oval shaped nodes are continuous variables with conditional log Gaussian distributions. The upper bar chart (2E) shows the prior waste distribution from dump records. The middle chart (2F) shows the posterior waste composition (WC) after updating the network with the assay results (ACD). Lower chart (2G) and displayed network shows posterior distributions given the concentrator reading of a soil sample (SS & ACD). The two graphs show predicted sample contamination SCD given assay results (ACD) before (2H) and after (2J) observation of sample sensor (SS). At the concentrator (fig 2D), the predicted soil sample contamination and sensor are shown. Even though the sensor reading is quite high (8.0), the model predicts a low contamination (-34.5). The reason is that the high sensor reading is an indicator of high amounts of the wt-0 waste type, which has a high ratio of masking agent to contaminate. The nodes WC, L, M, F are the nearest ancestral discrete nodes to SCD and are, therefore, the source of the mixture distributions of SCD.

---

2. Volume reduction can only be achieved with a small sort sample. TRU waste contamination is attached to soil particles. The contaminated particles could be uniformly distributed through the soil box but most likely the TRU waste is highly concentrated in small very active particles distributed irregularly in the box. The soil box assay only provides a total contamination measurement. Slag volume reduction is dependent on the ability to pick out the small samples of soil that contain the highly active particles.

3. Studies showed that the neutron sensor could not be effective with small samples and does not have the throughput required. The gamma sensor can achieve high throughput with small samples, but is biased as explained in the previous section.

4. Uncertainty arises because of the unknown state of the buried waste, the location of various waste types in the pit, the state of the containers, amount of leakage and the transport mechanisms for penetration of waste into the surrounding soil. These factors contribute to the contamination, dispersion and mixture composition of the soil input at the segmented sorter

A Bayesian network representing variables involved in in-situ leakage and transport and in-process assay and concentration of mixed waste was constructed from expert knowledge and a representative network is shown in Figure 2. This network can be partitioned into four components.

The waste characteristics component (2A) begins with the root node (WC) representing the waste composition of a section of the pit from which a soil batch was extracted. The prior is assessed from a three-dimensional database of the pit. The soil surrounding the waste is collected into large bins (100 cu ft). Each bin is a batch of soil to be processed by the concentrator. The prior distribution of WC is assessed from dump records and is shown in the upper bar chart. The column of nodes represents various characteristics of the waste types. For display purposes, the charts show 5 waste types and 7 characteristics. In a typical application there would be several times those numbers.

The container condition sub-network (2B) represents expert knowledge about the processes that can destroy container integrity and transport waste contamination into the surrounding soil. The potential for container leakage is affected both by the type of waste stored in the container and unknown events that may have occurred at the time of dumping up until the time of retrieval.

The average contamination of the batch is measured at an assay station. The assay sub-network (2C) predicts the assay results (ACD and AMD). The actual assay is asserted into the Bayesian network and the network is updated. This results in a conditional posterior distribution over the mixtures of waste (WC) in the batch (middle bar chart – 2F), as well as the distribution of the sample contamination (SCD) to be seen at the concentrator (graph 2H). Finally the concentrator sub-network predicts the radiation as measured by the concentrator sensor array (SS) as well as the actual contamination (SCD). As the batch is processed, the sensor array assesses the contamination of small soil samples (0.1 cu ft). The small sample sensor value is input to the Bayes network for another update (chart 2G). This second update provides a prediction of the actual sample contamination. (graph 2J) This prediction is the posterior probability distribution that is conditioned on the section of the pit where the batch was extracted, the information collected at the batch assay, the measured contamination of soil in the batch already processed, and the sensor reading for the current sample.

From this derived posterior of the actual sample contamination the expected loss of diverting the sample to the slag stream is compared with the expected loss of accepting it into the clean soil stream. The controller chooses the action that minimizes the expected loss.

## 3. A DECISION RULE FOR OPTIMAL CONTROL.

There are two actions for the controller: divert the sample to the slag stream or accept it as clean for the restoration of the pit. For simplicity, we assume that there is a rejection threshold $\hat{c}$ such that if a sample is accepted as clean when in fact its contamination is greater than $\hat{c}$ then a loss is incurred. Such a loss may be a penalty that is assessed or it may be the cost of reworking the sample when discovered at a later point in the process. Similarly, rejecting a sample that is clean also incurs a loss, for example, the cost of unnecessary maintenance and storage. This loss matrix is displayed in Table 1.

A decision rule D maps information to action. The optimal decision minimizes expected loss given the information:

$$D^*(I) = \arg\min \int_c \Lambda(d, c) \, dP(c \mid I)$$

It is not hard to show that $D^*$ has the following form when $\Lambda$ is given by table 1:

$$D^*(I) = \text{divert iff } P(c > \hat{c} \mid I) > \Lambda_0 / \Lambda_1.$$

| Table 1: $\Lambda(d, c)$ = Loss matrix of the divert decision | | |
|---|---|---|
| decision | outcome | |
|  | $c \leq \hat{c}$ | $c > \hat{c}$ |
| divert | $\Lambda_0$ | $\Lambda_0$ |
| do not divert | 0 | $\Lambda_1$ |
| $c$ = sample contaminate density | | |
| $\Lambda_0$ = cost of conversion to slag and long term storage | | |
| $\Lambda_1$ = cost of making the error of allowing a toxic sample to return to pit | | |

## 4. COMPILATION FOR A REAL TIME CONTROLLER

In theory, the Bayesian network of Figure 2, after updating with the assay and sensor observations, provides the distribution $P(c|I) = P(SCD \mid ACD, SS)$ needed for selecting the best control decision. However, there are two practical complications.

1. The network contains a mixture of discrete and continuous variables.
2. Solving the network for each new value of the sensor requires too much computing resources for real time control.

### 4.1 COMPLICATIONS IN A CG-NETWORK

The mixture of discrete and continuous variables can be handled using the cg-network algorithm proposed by Lauritzen [1992]. However, the Lauritzen algorithm only provides an approximation, the closest Gaussian to the mixture. This approximation can be very gross. Also, the algorithm needs some minor alteration to accommodate near zero values. It is not hard to modify the algorithm to accommodate these problems (see appendix). However, the time for updating the full network did not meet the requirements of the controller.

### 4.2 A REAL TIME ALGORITHM FOR BATCH PROCESSING

A Bayesian network is a representation of the joint distribution of the variables in a problem. Many variables that are needed to assess the network are unobserved, hidden variables that give structure and meaning to the network. Among the observed variables there are differences in the frequency of observation. Such variations can be exploited in the control setting for efficient processing. Controls are determined at run time only by the most frequently observed variables. Furthermore, only a few unobserved variables have a direct impact on the control decision. Thus, by eliminating most variables through instantiation (of the less frequently observed variables) or integration (of the hidden variables), a large Bayesian network can often be reduced to a small network of only a few variables. This is the strategy employed in hierarchical control (Sethi and Zhang, 1994). The human nervous system is an example of a hierarchical control system.

The optimal control requires $P(c \mid I)$, or $P(SCD \mid ACD, SS)$ in our example. At run time while processing a soil batch, the assay variable is fixed, so that what is required is $P(SCD \mid SS; ACD)$, the probability of the actual contamination given the observed measured value, holding ACD constant. In the segmented sorter controller, the network of figure 2 is reduced to that of figure 3 between the time a soil box leaves the soil assay station and is loaded into the sorter.

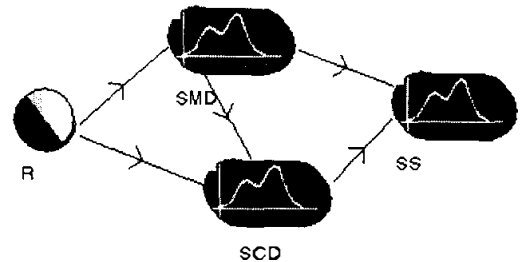

Figure 3: Run-time network compiled from network in figure 2.

The joint distribution of figure 3 is fairly easily derived from the clique tree representation of figure 2 to obtain the clique tree for figure 3. The runtime network (figure 3) is updated with each sensor value and the resulting distribution of SCD is used to compute the optimal control decision as given above in section 3.

When the variable SMD is also integrated out of the network of figure 3, the result is a mixture bivariate distribution for SCD and SS. (Figure 4). Each component of the mixture has a positively sloped major axis, indicating a positive correlation between sensor and sample contamination for a given waste type. However, the bivariate

Gaussian that minimizes the Kullback-Liebler distance to this mixture has a negatively sloped major axis, indicating an inverse relation between sensor and contamination. This is a result of the high gamma, low neutron radiation from the waste type wt-0. Such waste has relatively short life and should not be included in long term slag storage and maintenance.

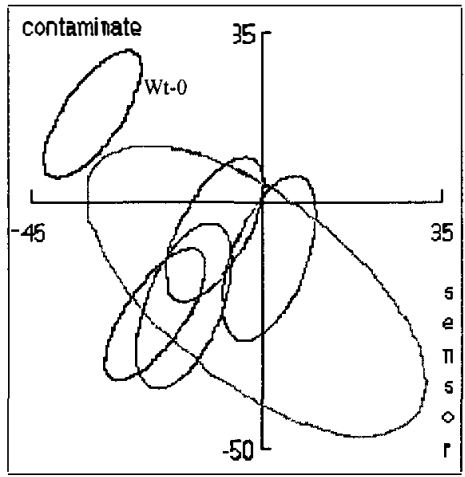

Figure 4. 95% ellipsoids of mixture components of the joint distribution of sensor (SS) and sample contaminate (SCD) given the assay reading. Dark lines indicate the components of the mixture distribution. The gray ellipsoid is the 95% probability level for the Gaussian approximation given by the Lauritzen algorithm.

A network like figure 3 is small enough that the network can be evaluated and a decision computed in less than a second, even with many waste types. If millisecond control decisions are required, the control decision rule can be computed as a function of the sensor reading, SS. The form of the rule will, in general, be a collection of real intervals $\{[s_k, s_{k+1}]\}$ in which diversion of the sample to the slag stream is optimal. This rule is simple enough that it can be loaded onto the PLC (programmable logic controller) implementing the divert action.

## 5. CONCLUSIONS

Many control problems involve uncertainty about the meaning of sensory information. Bayesian networks provide the analysis upon which an optimal control solution can be calculated. The computational complexity that comes with the Bayesian method can be an obstacle to implementing a Bayesian network based controller. Nevertheless, careful architectural design can reduce the resource requirements of runtime computation. In the example studied in this paper, much of the computation can be completed off line, during the staging of batches in a batch process system. Essentially one recognizes that many of the variables in the original network, though important for the original assessment, can be absorbed into a marginal distribution of variables that are either sensors that change during processing or key variables that are essential for making the control decision.

Computation of the marginal joint distribution can be computed through a mixture of strategies. Often this can be accomplished by using the potentials of the cliques for the larger network. Adding arcs to the network or constraints on the triangulation of the network can help. Conditioning methods as in footnote 1, and node elimination methods as found in the Shachter algorithm can also be employed.

A batch processing system is the simplest example where differences in the observation frequency provide a strategy for reducing the runtime computation of controls. Further study of network exact solution algorithms that are built for hierarchical control architectures holds promise for bringing the power of Bayesian networks to real time process control in highly uncertain environments.

### Acknowledgements

This proposed Bayesian control algorithm is in part based on a general problem identified by the authors while consulting for Lockheed Martin Advanced Environmental Systems on a waste remediation project at the Idaho National Environmental and Engineering Laboratory site. We greatly appreciate the opportunity provided by Lockheed. The application was prototyped using Bayes On-Line™ from Gensym corporation.

### Appendix

The following algorithm will yield the exact distribution of a continuous node X in a cg-network conditioned on a set of observed nodes, Z.

1. Update the network using Lauritzen's algorithm.

2. In the reversed arc direction find the collection of ancestral discrete nodes, $\{\delta_1 \ldots \delta_m\}$ nearest to X.

3. Set $\Pi = P(\delta_1 \mid Z)$ as given by the solution in (1) for a given value of $\delta_1$. Then instantiate $\delta_1$ to that value.

4. Recursively multiply the product $\Pi$ by $P(\delta_i \mid \delta_{i-1} \ldots \delta_1, Z)$ for some value of $\delta_i$. Then instantiate $\delta_i$ to that value.

5. Repeat 4 until $d_m$ is reached. $\Pi$ is the joint probability of all the possible sources of mixtures for X. Record the mixture probability $\Pi$ and the mean and variance of X.

6. When the recursion completes, the exact mixture distribution for X given Z has been obtained from the cg-network.

The repetitious updating of the network can be restricted to X and the continuous ancestors of X between X and $\{\delta_1 \ldots \delta_m\}$. This is often contained in a branch of the clique tree. Consequently, the updating in (4) need only be performed within this branch of the clique tree.

## References

Cooper, G.F., 1990. The computational complexity of probabilistic inference in Bayesian networks is NP-hard. *Artificial Intelligence*, 42(2-3), 393-405.

Dean, T., and Boddy, M., 1994. An analysis of time dependent planning, in *Proceedings of 7$^{th}$ National Conference on Artificial Intelligence*.

Horsch, M.C., and Poole, D. 1998. An anytime algorithm for decision making under uncertainty, in *Proceedings of the 14$^{th}$ Conference on Uncertainty in Artificial Intelligence*, pp 246 - 255.

Horvitz, E., 1987, Reasoning about beliefs and actions under computational resource constraints, in *Proceedings of 3$^{rd}$ Conference on Uncertainty in Artificial Intelligence*, pp 301-324.

Ibarguengoytia, P.H., Sucar, L.E., and Vadera, S., 1998, Any time probabilistic reasoning for sensor validation, *Proceedings of the 14$^{th}$ Conference on Uncertainty in Artificial Intelligence*, pp 266 - 273.

Lauritzen, S., 1992, Propagation of probabilities, means, and variances in mixed graphical association models, *Journal Of The American Statistical Association*, pp 1098-1108.

Shachter, R.D., 1986, Evaluating influence diagrams, *Operations Research*, 34(6) pp 871-882.

Sethi, S.P. and Zhang, Q., [1994] Hierarchical Decision Making in Stochastic Manufacturing Systems, Birkhauser